\pdfoutput=1

\documentclass[11pt]{article}

\usepackage[preprint]{acl}

\usepackage{times}

\usepackage{amsmath,amsfonts,bm}

\def\eqref#1{equation~\ref{#1}}

\def\1{\bm{1}}

\DeclareMathAlphabet{\mathsfit}{\encodingdefault}{\sfdefault}{m}{sl}
\SetMathAlphabet{\mathsfit}{bold}{\encodingdefault}{\sfdefault}{bx}{n}

\usepackage{tabularx}
\usepackage{hyperref}
\usepackage{url}
\usepackage{booktabs}
\usepackage{makecell}
\usepackage{multirow}
\usepackage{minitoc}
\usepackage{tikz}
\usetikzlibrary{shapes.geometric, arrows, positioning}

\tikzstyle{block} = [rectangle, draw, fill=blue!20, text width=8em, text centered, rounded corners, minimum height=4em]
\tikzstyle{line} = [draw, -latex']
\tikzstyle{cloud} = [draw, ellipse,fill=red!20, minimum height=2em]

\usepackage{CJKutf8}


\usepackage{amsmath}
\usepackage{algpseudocode}
\usepackage{amsfonts}
\usepackage{comment}
\usepackage[utf8]{inputenc}
\usepackage[linesnumbered,ruled,vlined]{algorithm2e}

\usepackage{ifthen}
\newboolean{showcomments}
\setboolean{showcomments}{true} 

\newcommand{\jianbo}[1]{\ifthenelse{\boolean{showcomments}}{\textcolor{purple}{[#1 - jianbo]}}{}}
\definecolor{pink}{RGB}{255,170,182}

\usepackage{multirow}
\usepackage{array}
\usepackage{graphicx}
\usepackage{diagbox}
\usepackage{tocloft} 
\usepackage{multicol} 
\usepackage{appendix}
\usepackage{tocloft}

\author{
Di Zhang$^{1,2}$\thanks{These authors contributed equally.}, Jianbo Wu$^{3*}$, Jingdi Lei$^{2*}$, Tong Che$^{4*}$ \\
\textbf{Jiatong Li}$^{5}$\textbf{,} \textbf{Tong Xie}$^{6}$\textbf{,} \textbf{Xiaoshui Huang}$^{7}$\textbf{,} \textbf{Shufei Zhang}$^{2}$\textbf{,} \textbf{Marco Pavone}$^{8}$\\
\textbf{Yuqiang Li}$^{2}$\thanks{Corresponding author} \textbf{,} \textbf{Wanli Ouyang}$^{2}$\textbf{,} \textbf{Dongzhan Zhou}$^{2\dag}$  \\
$^{1}$Fudan University, $^{2}$Shanghai Artificial Intelligence Laboratory, $^{3}$University of California, Merced\\ $^{4}$Independent Researcher, $^{5}$Hong Kong Polytechnic University, $^{6}$University of New South Wales\\ $^{7}$Shanghai Jiao Tong University, $^{8}$Stanford University\\
\texttt{\{liyuqiang,zhoudongzhan\}@pjlab.org.cn} 
}

\usepackage{todonotes}

\usepackage{latexsym}

\usepackage[T1]{fontenc}

\usepackage[utf8]{inputenc}

\usepackage{microtype}

\usepackage{inconsolata}

\usepackage{graphicx}

\title{LLaMA-Berry: Pairwise Optimization for Olympiad-level Mathematical Reasoning via O1-like Monte Carlo Tree Search}

\renewcommand{\tableofcontents}{\section*{Contents}\cftsetindents{section}{0em}{2em}}

\begin{document}

\maketitle

\begin{abstract}
This paper presents an advanced mathematical reasoning framework, \texttt{LLaMA-Berry}, for enhancing the problem-solving ability of large language models~(LLMs). The framework combines Monte Carlo Tree Search with Self-Refine (SR-MCTS) to optimize the reasoning paths and utilizes a pairwise reward model to evaluate different paths globally. By leveraging the self-critique and rewriting capabilities of LLMs, our SR-MCTS overcomes the inefficiencies and limitations of conventional step-wise and greedy search algorithms by fostering a more efficient exploration of solution spaces. To guide the search process, we propose Pairwise Preference Reward Model~(PPRM) to predict pairwise preferences between solutions through instruction-following capabilities trained by Reinforcement Learning from Human Feedback (RLHF). Finally, the Enhanced Borda Count (EBC) method is adopted to synthesize pairwise preferences into global quantile scores for evaluations. This approach addresses the challenges of scoring variability and non-independent distributions in mathematical reasoning tasks. The framework has been tested on general and advanced benchmarks, showing superior search efficiency and performance compared to existing open-source and closed-source methods, particularly in complex Olympiad-level benchmarks, including AIME24 and AMC23.
\end{abstract}



\section{Introduction}

\begin{figure*}[!htbp]
    \centering
    \includegraphics[width=0.94\linewidth]{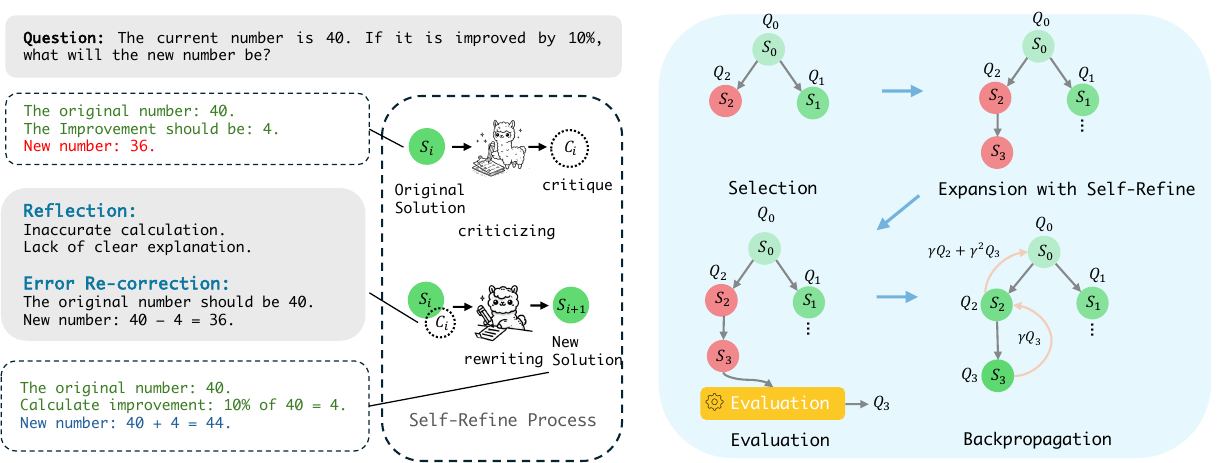}
    \caption{The main pipeline of \texttt{LLaMA-Berry}, where $S_i$ stand for problem-solving solutions and $C_i$ stands for critiques. The pipeline consists of four phases detailed in Section~\ref{sec:Self-refine Monte Carlo Tree Search}, including selection, expansion, evaluation, and backpropagation.}
    \label{fig:overview}
\end{figure*}
Mathematical reasoning represents a great challenge in artificial intelligence, with broad applications across automated theorem proving, mathematical problem solving, and scientific discovery~\citep{ahn2024large}. Recently, significant strides have been made by large language models (LLMs) like GPT-4~\citep{achiam2023gpt} in general mathematical tasks involving arithmetic and geometric problem-solving~\citep{cobbe2021training, Easy-to-Hard-Generalization, ying2024internlmmath}. 
However, complex mathematical reasoning remains challenging, especially at the Olympiad-level benchmarks such as AIME~\citep{AIME}. %

An intuitive approach to improving problem-solving is to break solutions into step-by-step reasoning paths~\citep{lightman2023let, luo2024improvemathematicalreasoninglanguage}, as demonstrated in Chain-of-Thought (CoT~\citealp{wei2022chain}). While prompt-based methods can effectively facilitate the construction of such reasoning paths, they may still encounter challenges due to the lack of comprehensive feedback during the generation process, which can affect efficiency~\citep{paul2023refiner}. In contrast to stepwise generation methods, another promising line of research treats the entire solution as an independent state, employing rewriting capabilities to refine the solutions, such as in Self-Refine~\citep{Self-Refine-first} and Reflexion~\citep{Reflexion}. However, these approaches, while innovative, may occasionally face challenges like being susceptible to local optima or potentially drifting towards suboptimal solutions due to flawed feedback, which could impact their maximum potential performance.

In addition to generating reasoning paths, effective solution evaluation is crucial, with models like the outcome reward model (ORM) and process reward model (PRM)~\citep{uesato2022solving} serving as valuable examples. The ORM focuses on the correctness of the final answer in a reasoning path, while the PRM emphasizes the correctness of each step in the process. While both methods enable reward models to assign scalar scores, obtaining reliable labeled data for training these reward models remains a significant challenge. Moreover, the scoring standards for mathematical reasoning tasks can vary significantly, as each problem presents unique characteristics. This variation complicates the scaling of reward models and hinders their ability to capture local preference relations between solutions. Although trained using language models, these reward models have yet to fully leverage instruction-following capabilities, which may limit their effectiveness in handling more complex reasoning tasks~\citep{genRM}.

To improve the efficiency of solution search in mathematical problems, we treat a complete solution as an independent state and apply Self-Refine to optimize previous solutions in order to obtain better ones. In the Self-Refine process, feedback from critiques is utilized to make the search more efficient compared to stepwise reasoning path generation. Furthermore, we incorporate Monte Carlo Tree Search (MCTS~\citealp{kocsis2006bandit}) to replace the iterative manner in naive Self-Refine, enhancing the solution search. MCTS leverages signals from the evaluation process to assess solutions and uses the Upper Confidence Bound applied to Trees~(UCT) method to balance exploration and exploitation. This approach enables the search process to effectively exploit higher-quality solutions and explore those with greater potential for improvement, while avoiding getting trapped in suboptimal local minima. 

In evaluation process, utilizing the instruction-following capabilities trained by Reinforcement Learning from Human Feedback~(RLHF~\citealp{christiano2017deep}), Pairwise Preference Reward Model~(PPRM) transforms the absolute rewards calculation into preferences prediction between solutions to calculate rewards. The approach reduces the variability with scoring characteristics and thus leads to a more robust and consistent evaluation of different solutions. To overcome the locality limitations inherent in pairwise comparisons, we employ the Enhanced Borda Count (EBC) method to aggregate local preference evaluations into global quantile scores, leading to more informed decision-making and, ultimately, better solutions. Combining the PPRM and EBC method not only enables the reward model to learn a more robust reward signal but also captures the global characteristics of the solution space, ensuring more reliable comparisons.

Our contributions are summarized as follows: (1) We propose \textbf{SR-MCTS}, a novel Markov Decision Process (MDP) framework that treats entire solutions as states and Self-Refine as optimization action to perform advanced solution search with MCTS. (2) \textbf{PPRM} is developed to leverage the preference relationship between solutions to evaluate their quality, which avoids the volatility of absolute scores while providing a more guided exploration of optimal paths. We adopt the EBC method to convert the local preferences into global evaluations. (3) We verify the effectiveness of \texttt{LLaMA-Berry} on multiple benchmarks, which outperforms baseline approaches like ToT~\citep{yao2024tree} and rStar~\citep{Mutual-Reasoning} in both search efficiency and accuracy. Notably, \texttt{LLaMA-Berry} enhances the performance of LLaMA-3.1-8B, making the 8B-level model comparable to proprietary models, including GPT-4 Turbo on Olympiad-level mathematical reasoning \textbf{without additional training}.

\section{Methodology}
\subsection{Preliminary}
One of the core challenges in mathematical problem-solving is to generate and optimize reasoning paths to derive high-quality solutions. We formalize this process in a path-wise Markov Decision Process (MDP) framework, where each state \( s \) in the state space $S$ represents a \textit{complete solution} to a given problem, and the action space \( A \) consists of all feasible \textit{rewriting} actions $a$ that make transitions between states.


\begin{figure*}[!tbp]
    \centering
    \includegraphics[width=1.0\linewidth]{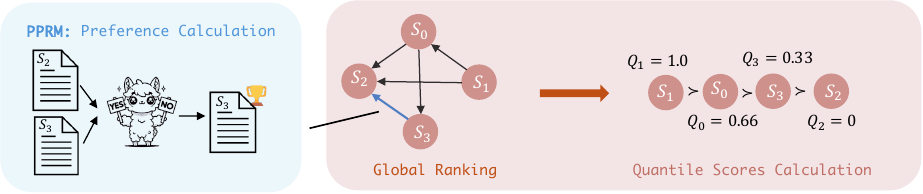}
    \caption{Preference prediction process of PPRM and global quantile score based on Enhanced Borda Count method.}
    \label{fig:self-refine}
\end{figure*}

In the framework, we aim to quantify the expected reward \( Q(s, a) \) from executing action \( a \) at state \( s \), that is,

\begin{equation} 
Q(s, a) = \mathbb{E}[R(s') | s' = T(s, a)]
\end{equation}

\noindent, where $T(s,a)$ indicates the transition from $s$ to another solution $s'$ via the rewriting action $a$. Our primary objective is to identify the optimal state \( s^* \) that represents the best solution. We can reach \( s^* \) by selecting actions that maximize the reward, guiding us toward the most desirable outcome, as demonstrated in Equation~\ref{eq:optimal}.

\begin{equation}
\label{eq:optimal}
s^* = \arg\max_{s' \in S} Q(s')
\end{equation}



\subsection{Self-Refine applied to MCTS}
\label{sec:Self-refine Monte Carlo Tree Search}

As shown in Figure~\ref{fig:overview}, SR-MCTS integrates Monte Carlo Tree Search (MCTS) with the Self-Refine mechanism to continuously evaluate and optimize the solution search. This integration leverages the iterative nature of MCTS and the self-improvement capabilities of LLMs, thereby improving the search outcomes.

Monte Carlo Tree Search (MCTS) is an effective method within the Markov Decision Processes (MDP) framework, employing states, actions, and value functions through sampling. The algorithm follows four key steps: selection, expansion, evaluation, and backpropagation. In the selection phase, the root node is expanded using the Upper Confidence Bound applied to Trees (UCT) algorithm, which selects a node \( s \) by balancing exploration and exploitation:

\begin{equation} 
a = \arg\max_{a \in A(s)} \left(Q(s, a) + c \cdot \sqrt{\frac{\ln N(s)}{N(s, a)}} \right)
\end{equation}

\noindent, where \( N(s) \) is the visitation count of node \( s \), \( N(s, a) \) is the action frequency, and \( c \) is a parameter controlling exploration. In the expansion phase, node \( s \) generates subsequent states \( s' \), added as new nodes in the tree \( \mathcal{T} \). The evaluation phase typically uses simulations or heuristics to estimate the Q-values for these nodes. Finally, during backpropagation, the estimated Q-values are updated retroactively from the leaf nodes to the root. This iterative process allows MCTS to refine decision-making by balancing the exploration of new paths with the exploitation of known high-value paths.


\noindent\textbf{Selection phase.} The selection phase identifies a node \( s_i \) from the search tree \( \mathcal{T} \) for expansion, where each node represents a complete solution state. The Upper Confidence Bound applied to Trees~(UCT) algorithm is employed to select the optimal node, with dynamic pruning used to avoid local optima. A node \( s_i \) is considered fully explored when its child nodes reach a predefined limit, and at least one child node's Q value exceeds the Q value of \( s_i \). 

\noindent\textbf{Expansion phase.} In the expansion phase, as shown in Figure~\ref{fig:overview}, the selected answer \( s_i \) is expanded by generating successor answers through a Self-Refine process, which includes a \textbf{Criticizing} and \textbf{Rewriting} process. The \textbf{Criticizing} process generates a critique \( c_i = C(s_i) \) that identifies drawbacks (e.g., mathematical wrongs or logical faults) in the current chosen answer $S_i$, and then \textbf{Rewriting} process generates a new answer \( s_{i+1} = R(s_i, c_i) \). In practice, to simplify the problem, we assume this process is deterministic, ensuring that the same original state of solutions $s_i$ consistently produces the same successor state of solution \( s_{i+1} \). The new state of solution \( s' \) is then added to the search tree \( \mathcal{T} \) as a new node.

\noindent\textbf{Evaluation phase.} The evaluation phase calculates the value \( Q(s') \) of the newly generated node \( s' \) using the Pairwise Preference Reward Model~(PPRM). The evaluation involves two steps: global and local value assessments. The global value \( Q_g(s') \) is determined by the quantile of \( s' \) in a win-loss preference matrix \( \mathbf{M} \), which reflects the win-loss relationships between nodes. The local value \( Q_l(s') \) is derived from comparisons with adjacent nodes in the search tree \( \mathcal{T} \). The total value \( Q(s') \) is then computed as a weighted combination of global and local values: \( Q(s') = \alpha Q_g(s') + (1 - \alpha) Q_l(s') \), where \( \alpha \) controls the relative influence of each component.

\noindent\textbf{Backpropagation phase.} In the backpropagation phase, the value \( Q(s') \) of the new node is propagated back to its parent node \( s_i \), updating Q value of \( s_i \) as a discounted sum of its child nodes' Q values: \( Q(s_i) = (1 - \gamma) Q(s_i) + \gamma Q(s') \). The discount factor \( \gamma \) represents the importance of future rewards. This iterative update mechanism ensures that the values of parent nodes are progressively refined, enhancing the guidance for future selections.

Additionally, to control the growth of the search tree, the SR-MCTS method restricts the maximum number of rollouts \( N_{max} \) . The search process terminates when the restriction is reached, imposing limits on the unbounded expansion of the tree. The overarching objective of SR-MCTS is to maximize the expected highest $Q$ value of all existing nodes $S$, guiding us towards the most desirable outcome $s^*$, ensuring that the search process efficiently converges to high-quality solutions.

\subsection{Pairwise Preference Reward Model}

Reliable evaluation of different solutions plays a crucial role in mathematical problem-solving tasks as it leads to better estimation of Q-values, thereby offering better guidance. Existing reward models typically evaluate solutions by giving absolute scores, such as process reward model~(PRM~\citealp{lightman2023let}) and outcome reward model~(ORM~\citealp{yu2023outcome}). However, the score-based reward models may fall short in leveraging the instruction-following capability of LLMs or effectively handling the variations in scoring standards, especially when the differences between solutions are subtle.
To address this issue, we propose the Pairwise Preference Reward Model~(PPRM), which leverages a comprehensive preference dataset incorporating substantial samples from both PRM and ORM approaches~\citep{OpenMathInstruct-1,lightman2023let} to learn preference relationships among mathematical solutions. 

For two solutions ($a_1$ and $a_2$) to a given mathematical problem, we use $a_1 \succ a_2$ to represent the situation where $a_1$ is preferred over $a_2$. PPRM predicts their relative quality using a pairwise partial ordering, represented by the following probability formula:

\begin{equation}
P(a_1 \succ a_2 \mid \phi) = \frac{e^{\phi(a_1)}}{e^{\phi(a_1)} + e^{\phi(a_2)}}
\end{equation}

\noindent, where $P(a_1 \succ a_2 \mid \phi)$ denotes the probability of a partial ordering relation between solution $a_1$ and $a_2$, with \(\phi\) representing the parameters of the reward model. In our method, $a_1 \succ a_2$ are represented by tokens of an LLM, and $P(a_1 \succ a_2 \mid \phi)$ is estimated using the logits value of tokens calculated by the LLM.

Then, inspired by advancements in Language Interface Fine-Tuning (LIFT~\citealp{dinh2022lift}), we frame the training process of PPRM as a question-answering task to leverage the instruction-following capabilities of LLMs. The model is tasked with answering the question, "For Question \(Q\), is solution \(a_1\) better than solution \(a_2\)?" as shown in Figure~\ref{fig:self-refine}. To form a robust training objective, the predicted token labels \(\hat{y}\) ('Yes' or 'No') are evaluated with ground truth label $y$ using the indicator function \(\mathbf{I}\):

\begin{equation}
\mathbf{I}(\hat{y}, y) = 
\begin{cases} 
1 ,& \text{if } \hat{y} = y \\
0 ,& \text{if } \hat{y} \neq y 
\end{cases}
\end{equation}

Finally, a pairwise preference dataset \(\mathcal{D}\) that contains millions of mathematical problem-solving solution pairs is converted into a dataset \(\mathcal{D}'\) suitable for a question-answering task. We employ RLHF techniques to train the model to improve its performance in the partial-order prediction question-answering task. Subsequently, the Direct Preference Optimization (DPO~\citealp{rafailov2024direct}) method is utilized to find the optimal \(P_\phi\) by maximizing the objective \(\arg\max_{\phi} \mathbb{E}_{P} [\mathbf{I}(\hat{y}, y)]\). Please refer to Appendix~\ref{sec:Details of PPRM Trainging} for details about training and inference of PPRM.

\subsection{Enhanced Borda Count Method}

Although PPRM allows us to directly compare the quality of two solutions, we still need to convert these local preferences into a cohesive global ranking to gain a comprehensive evaluation for the answers. 
This conversion process can be formalized as the global optimal ranking aggregation~(GORA) problem related to Learning to Rank~(LtR) methods. Further, we propose the Enhanced Borda Count~(EBC) method based on the transitivity assumption of mathematical problem solutions, which integrates the naive Borda Count algorithm with a transitive closure of preferences calculated by the Floyd-Warshall~\citep{10.1145/321105.321107} algorithm. For formalized discussion, please refer to Appendix~\ref{sec:Proof of Convergence of the Enhanced Borda Count (EBC) Method}.

\noindent\textbf{Local preference calculation.} First, the PPRM generates a preference matrix \(\mathbf{M} \in \mathbb{R}^{n \times n}\) for all $n$ problem solutions, where \(\mathbf{M}_{i,j} = 1\) indicates that solution \(a_i\) is superior to solution \(a_j\), and \(\mathbf{M}_{i,j} = 0\) otherwise. This process can be represented as:
    
\begin{equation}
\mathbf{M}_{i,j} = 
\begin{cases}
1, & \text{if } P(a_i \succ a_j) \geq 0.5 \\
0, & \text{if } P(a_i \succ a_j) < 0.5
\end{cases}
\end{equation}

\noindent As shown in Figure~\ref{fig:self-refine}, this matrix can be viewed as an adjacency matrix of a directed graph \(\mathbf{G} = (\mathbf{V}, \mathbf{E})\), where each solution \(a_i\) corresponds to a vertex \(v_i\), and an edge \(e = (v_i, v_j)\) exists if \(\mathbf{M}_{i,j} = 1\), indicating that solution \(a_i\) is preferred over \(a_j\).

\noindent\textbf{Transitive closure.} To simplify the problem, we adopt the assumption of transitivity for mathematical solutions, that is, if \(v_i \succ v_k\) and \(v_k \succ v_j\), then \(v_i \succ v_j\). Under this assumption, the transitive closure \(\mathbf{C}\) of a preference matrix can be computed through Floyd-Warshall algorithm, e.g., if \(\mathbf{M}_{i,k} = 1\) and \(\mathbf{M}_{k,j} = 1\), then \(\mathbf{M}_{i,j} = 1\). 



\noindent\textbf{Borda count-based global ranking.} Next, based on the transitive closure matrix \(\mathbf{C}\), we apply the Enhanced Borda Count method for global ranking. The Enhanced Borda Count determines the ranking of each node by calculating its out-degree, which corresponds to the number of nodes it defeats. For each node \(v_i\), the $\text{Borda}(v_i)$ is defined as $\sum_{j=1}^{n} \mathbf{C}_{i,j}$, like the ranked node listed in Figure~\ref{fig:self-refine}.

Nodes with higher Borda counts are ranked higher. However, in practice, cyclic preferences can cause efficiency issues with the naive Borda Count method. To further refine the ranking, we devise a re-ranking stage, where the logits generated by the PPRM are used for soft comparison among nodes with equal Borda counts. Specifically, for two nodes \(v_i\) and \(v_j\) with equal Borda counts, the soft comparison rule can be denoted as $v_i \succ v_j \iff P(a_i \succ a_j) > P(a_j \succ a_i)$. This process ensures that the ranking remains consistent and reasonable even in the presence of cycles or local ambiguities.

\noindent\textbf{Global quantile score of solutions.} Finally, the ranking is converted into a global quantile score \( Q_g \) for each solution \(v\) is \(Q_g(v) = 1 - \frac{\text{rank}(v)-1}{|V|-1}\), where \(\text{rank}(v)\) is the position of \(v\) in the ranking based on Borda counts, and \(|V|\) is the total number of nodes. To reflect local advantages in the search tree structure, the local win rate \(Q_l(v)\) for a node \(v\) is calculated in $\mathbf{C}$ with its children nodes $\text{Children}_{v}$ as follows:

\begin{equation}
Q_l[v] = \frac{\sum_{u \in \text{Children}[v]} C[u, v]}{|\text{Children}[v]|}
\end{equation}

Finally, score \(Q(v)\) for a solution is a weighted sum of local win rate \(Q_l(v)\) and global quantile score \( Q_g \). 




\begin{table*}[!tbp]
\centering
\caption{Performance comparison of models across benchmarks of different difficulties, as represented by GaoKao2023En~\citep{MARIO}, College Math~\citep{tang2024mathscale}, and OlympiadBench~\citep{he2024olympiadbench}, which range from high school to Olympiad levels. Scores denoted with subscripted notations, such as \texttt{maj@8}, represent specific metrics, with \textbf{major@8} as an example. Scores without subscripted notations reflect the model's greedy performance evaluated in a zero-shot CoT manner.}
\label{tab:math-models-benchmark}
\resizebox{2.\columnwidth}{!}{
\begin{tabular}{c|llllll}
\toprule
\hline
\diagbox{Model}{Benchmark}                                   & GSM8K & MATH & GaoKao2023En & OlympiadBench & College Math & MMLU STEM \\ \hline
Qwen2-7B-Instruct~\citep{qwen2}                              & 85.7  & 52.9 & 36.4           & 21.3          & 24.5         & 68.2      \\ 
Meta-Llama-3.1-8B-Instruct~\citep{meta_llama_8b_2024}        & 76.6  & 47.2 & 30.1           & 15.4          & 33.8         & 60.5      \\ \hline
Qwen2-72B-Instruct~\citep{qwen2}                             & 93.2  & 69.0 & 58.7           & 33.2          & 43.2         & 84.4      \\ 
Meta-Llama-3.1-70B-Instruct~\citep{meta_llama_70b_2024}      & 94.1  & 65.7 & 54.0           & 27.7          & 42.5         & 80.4      \\\hline
DeepSeekMath-7B-RL~\citep{shao2024deepseekmath}              & 88.2  & 52.4 & 43.6           & 19.0          & 37.5         & 64.8      \\ 
Internlm2-math-plus-7b~\citep{ying2024internlmmath}          & 84.0  & 54.4 & 50.1           & 18.8          & 36.2         & 55.2      \\ 
Mathstral-7B-v0.1~\citep{mistralai2024mathstral}             & 84.9  & 56.6 & 46.0           & 21.5          & 33.7         & 64.0      \\ 
NuminaMath-7B-CoT~\citep{numina_math_7b}                     & 75.4  & 55.2 & 47.5           & 19.9          & 36.9         & 60.8      \\ 
Qwen2-Math-7B-Instruct~\citep{qwen2}                         & 89.9  & 75.1 & 62.1           & 38.2          & 45.9         & 63.8      \\ \hline
NuminaMath-72B-CoT~\citep{numina_math_72b}                   & 90.8  & 66.7 & 58.4           & 32.6          & 39.7         & 64.5      \\ 
Qwen2-Math-72B-Instruct~\citep{qwen2}                        & 96.7  & 84.0 & 68.3           & 43.0          & 47.9         & 79.9      \\ 
\hline
\multirow{2}{*}{\textbf{\makecell{Meta-Llama-3.1-8B-Instruct~\citep{meta_llama_8b_2024} \\+ LLaMA-Berry~(Ours)@8}}} & $89.8_{\texttt{maj@8}}$ & $54.8_{\texttt{maj@8}}$ & $36.4_{\texttt{maj@8}}$ & $24.8_{\texttt{maj@8}}$ & $36.4_{\texttt{maj@8}}$ & $68.3_{\texttt{maj@8}}$\\ 
& $94.9_{\texttt{rm@8}}$ & $69.4_{\texttt{rm@8}}$ & $61.6_{\texttt{rm@8}}$ & $47.2_{\texttt{rm@8}}$ & $63.7_{\texttt{rm@8}}$         & $82.9_{\texttt{rm@8}}$      \\ 
\hline
\textbf{+LLaMA-Berry~(Ours)@16} & $96.1_{\texttt{rm@16}}$  & $75.3_{\texttt{rm@16}}$ & $68.6_{\texttt{rm@16}}$           & $55.1_{\texttt{rm@16}}$          & $68.9_{\texttt{rm@16}}$        & $88.3_{\texttt{rm@16}}$      \\ \hline \bottomrule
\end{tabular}}
\end{table*}

\section{Evaluation}
\label{sec:Evaluation}

\subsection{Experiment Settings}

\noindent\textbf{Settings.} To better evaluate the effectiveness of our approach, we select LLaMA-3.1-8B-Instruct model~\citep{meta_llama_8b_2024} as the base model for SR-MCTS, \textbf{without any additional training}. We also train a Gemma2-2B-Instruct model~\citep{gemma_2b_it} as PPRM to provide reward signals during the search process. We develop the \texttt{Berry-Tree} inference framework to ensure robust and efficient inference, which supports advanced features, including fault tolerance, checkpoint recovery, multi-query concurrency, and automatic multi-server load balancing. Hyper-parameter settings are detailed in Appendix~\ref{sec:Hyper-paramerer settings}.

\noindent\textbf{Grading.} We evaluate algorithm-generated answers using the correctness evaluation standards as in~\citet{lightman2023let}, focusing on format adherence and content accuracy. The model is provided with a prompt specifying the expected answer format. We score answers as consistent if they exactly match the ground truth, closely approximate it numerically, or are equivalent in symbolic form. To ensure a comprehensive and rigorous evaluation, we adopt \textbf{major@k}~\citep{kuncheva2014combining} and \textbf{rm@k}~\citep{yang2024qwen2}, which can be unified as the solved rate of problems~\citep{lightman2023let,luo2024improvemathematicalreasoninglanguage}. The evaluation methods and metric details are further outlined in Appendix~\ref{sec:grading and metrics}.

\subsection{Benchmarks}

\textbf{General mathematical reasoning benchmarks. } We summarize the results on general mathematical reasoning benchmarks in Table~\ref{tab:math-models-benchmark}, which indicates that our method significantly boosts the base model's performance. The results consistently demonstrate improvements across various levels of difficulty. Specifically, the solved rate of problems in 16 rollouts of Meta-Llama-3.1-8B-Instruct~\citep{meta_llama_8b_2024} has been improved by more than 35\% on three benchmarks. Qwen2-Math-72B-Instruct~\citep{yang2024qwen2technicalreport} exhibits the strongest mathematical reasoning capability among the competing methods, while our \texttt{LLaMA-Berry}, built on a base model with only 8B parameters, exceeds it in the solved rate of problems on four benchmarks. In particular, \texttt{LLaMA-Berry} reaches 55.1\% on OlympiadBench and 68.9\% on College Math, surpassing it by 11.9\% and 21\%, respectively. 

\noindent\textbf{Cutting-edge mathematical Olympiad benchmarks. } In Table~\ref{tab:olympic-dataset-comparison}, we compare the performance of \texttt{LLaMA-Berry} with other leading models on Olympic-level benchmarks. The results demonstrate that \texttt{LLaMA-Berry} is highly competitive on these benchmarks, demonstrating its capability in complex reasoning. Notably, on the most challenging AIME2024 benchmark, our method boosts the base model’s solving rate from 2/30 to 8/30, surpassing typical open-source models and commercial closed-source models, except for the OpenAI o1 series~\citep{openAI-o1}. 

In addition to excelling in mathematical reasoning, our approach also excels across various science and engineering domains. For example, it achieves top performance on benchmarks such as MMLU STEM~\citep{hendryckstest2021} in Table~\ref{tab:math-models-benchmark} and GPQA diamond~\citep{rein2024gpqa} in Table~\ref{tab:olympic-dataset-comparison}. This demonstrates the method's robustness and versatility, enabling it to tackle a wide range of technical challenges and highlighting its potential for broader applications in both research and practical scenarios.

\begin{table}[!htbp]
\centering
\caption{Performance comparison across multiple olympiad benchmarks, including AIME24~\citep{AIME}, AMC23~\citep{AMC}, Math Odyssey~\citep{fang2024mathodysseybenchmarkingmathematicalproblemsolving}, and GPQA Diamond~\citep{rein2024gpqa}.}
\label{tab:olympic-dataset-comparison}
\scriptsize
\setlength{\tabcolsep}{1pt} 
\resizebox{1.02\columnwidth}{!}{
\begin{tabular}{l|llll}
\toprule
\hline
\multirow{2}{*}{Model} & \multicolumn{4}{c}{Benchmarks} \\
\cline{2-5}
 & AIME24 & AMC23 & Math Odyssey & GPQA$_{\texttt{Diamond}}$ \\
\hline
Claude 3 Opus                       & 6.7  & 42.0   & 40.6 & 50.4      \\
GPT 4 Turbo                         & 3.3  & --   & 47.0 & 38.8       \\
GPT 4o                              & 13.4 & --   & --   & 56.1    \\
OpenAI o1 Preview                   & 56.7 & --   & --   & 78.3       \\
OpenAI o1                           & 83.3 & --   & --   & 78.0      \\
Gemini 1.5 Pro                      & 6.7  & --   & 45.0 &  --   \\
Gemini Math-Specialized 1.5 Pro     & 23.3 & --   & 55.8 & --      \\
Meta-LLaMA-3.1-8B-Instruct          & 6.7  & 15.7 & 41.7 & 30.4  \\
Qwen2-Math-7B-Instruct              & 13.3 & 62.5 & --   & --        \\
NuminaMath-72B CoT                  & 3.3  & 52.5 & --   & --       \\
Qwen2-Math-72B-Instruct             & 20.0 & 60.0 & --   & --     \\
\hline
\multirow{2}{*}{\textbf{\makecell{Meta-Llama-3.1-8B-Instruct\\+ LLaMA-Berry~(Ours)@8}}}
 & $13.3_{\texttt{maj@8}}$ & $22.9_{\texttt{maj@8}}$ & $44.2_{\texttt{maj@8}}$ & $39.4_{\texttt{maj@8}}$  \\
& $16.7_{\texttt{rm@8}}$ & $48.2_{\texttt{rm@8}}$ & $60.4_{\texttt{rm@8}}$ & $77.3_{\texttt{rm@8}}$  \\
\hline
\textbf{ +LLaMA-Berry~(Ours)@16} & $26.7_{\texttt{rm@16}}$ & $54.2_{\texttt{rm@16}}$ & $65.0_{\texttt{rm@16}}$ & $92.4_{\texttt{rm@16}}$  \\
\hline
\bottomrule
\end{tabular}
}
\end{table}

\noindent\textbf{Comparison with other tree-based or CoT methods. } We compare our algorithm with other tree-based reasoning methods and CoT-based methods on GSM8K~\cite{cobbe2021training}, GSMHard~\citep{gao2022pal}, and MATH500~\citep{lightman2023let} which is a representative and highly challenging 10\% subset of MATH~\cite{hendrycks2021measuring} benchmark. 
As shown in Table~\ref{tab:different-trees-benchmark}, the performance of RAP~\cite{hao2023reasoning} and ToT~\citep{ToT} tends to degrade relative to more straightforward methods like Few-shot CoT and One-turn Self-Refine when the difficulty increases from GSM8K to GSMHard. We suspect the reason could be the weak self-evaluation capability of LLMs, which may guide reasoning steps to the inefficient side. Moreover, tree-based methods can incur more computational overhead than straightforward methods. In contrast, rStar~\cite{Mutual-Reasoning} and our method maintain a positive output performance trend, highlighting both approaches' higher search efficiency.

To make a fair comparison between the reported results on LLaMA-3-8B-instruct from rStar~\cite{Mutual-Reasoning}, self-consistency~\citep{wang2022self}, and our algorithm, we also utilize the LLaMA-3-8B-instruct as the base model instead of the 3.1 version. We observe that our approach achieves on-par or even better performance with fewer rollouts. Specifically, our method achieves an accuracy of 88.1\%, 31.5\%, and 39.6\% on GSM8K, GSMHARD, and MATH500 benchmarks, respectively, using the majority voting metric, which is among the same accuracy level as others while only consuming 1/2 of the rollout times of rStar and 1/8 of Self-consistency. This provides compelling validation of the efficacy of the EBC method and the aggressive exploration fostered by the dynamic pruning strategy.

\begin{table}[!t]
\centering
\caption{Performance of different tree-based methods for LLaMA-3-8B-Instruct on GSM8K, GSMHARD, and MATH500 benchmarks.}
\label{tab:different-trees-benchmark}
\scriptsize
\setlength{\tabcolsep}{3.0pt} 
\resizebox{0.49\textwidth}{!}{
\begin{tabular}{l|lll}
\toprule
\hline
\multicolumn{1}{c|}{\diagbox{Method}{Benchmark}} & GSM8K& GSMHARD  & MATH500\\
\hline
Zero-Shot CoT                  & 68.4  & 14.9  & 5.8  \\ 
Few-shot CoT                   & 74.5  & 25.6  & 17.8 \\ 
One-turn Self-Refine           & 75.7  & 26.5  & 25.0 \\ 
Self-Cons@8& 78.4& 28.5& 30.0\\ 
Self-Cons@64 & 83.2& 30.3& 33.0\\ 
Self-Cons@128& 84.7& 31.2& 33.8\\ 
ToT@32               & 69.1  & 19.6  & 13.6 \\ 
RAP@32          & 80.6  & 29.6  & 18.8 \\
rStar@32          & $88.7_{\texttt{maj@32}}$ & $33.4_{\texttt{maj@32}}$ & $38.3_{\texttt{maj@32}}$ \\ 
\hline
\multirow{2}{*}{\textbf{\makecell{LLaMA-Berry~(Ours)@8}}}         & $86.4_{\texttt{maj@8}}$ & $30.2_{\texttt{maj@8}}$ & $35.2_{\texttt{maj@8}}$ \\
    & $94.1_{\texttt{rm@8}}$  & $37.1_{\texttt{rm@8}}$  & $56.4_{\texttt{rm@8}}$  \\ 
    \hline
\multirow{2}{*}{\textbf{LLaMA-Berry~(Ours)@16}}         & $88.1_{\texttt{maj@16}}$ & $31.5_{\texttt{maj@16}}$ & $39.6_{\texttt{maj@16}}$ \\
    & $96.4_{\texttt{rm@16}}$ & $41.1_{\texttt{rm@16}}$ & $63.8_{\texttt{rm@16}}$ \\
\hline
\bottomrule
\end{tabular}}
\end{table} 
\subsection{Ablation Study}

As shown in Table \ref{tab:performance-breakdown}, we conduct ablation experiments to evaluate the key components of \texttt{LLaMA-Berry}, using the solved rate of problems~\citep{luo2024improvemathematicalreasoninglanguage} as a metric across benchmarks of increasing difficulty: GSM8K and AIME2024. Zero-shot CoT represents the base model’s greedy mathematical reasoning capabilities. SR-MCTS without PPRM refers to a basic version of our method that uses self-evaluation as the reward instead of PPRM and EBC.

When comparing iterative Self-Refine with SR-MCTS, especially on the GSM8K dataset, the introduction of MCTS effectively mitigates the issue of solution degradation into suboptimal results caused by flawed critiques in iterative methods. With rollouts of 8 and 16, SR-MCTS improves the solved rate of problems by 3.1\% and 3.2\%, respectively. Furthermore, in comparing SR-MCTS without PPRM and SR-MCTS with PPRM, PPRM further boosts the solved rate of problems on the relatively easier GSM8K dataset. With rollouts of 8 and 16, the solved rate of problems is elevated by 13\% and 14.7\%, respectively.

\begin{table}[t]
\centering
\caption{Ablation study for \texttt{LLaMA-Berry} framework. The PPRM and MCTS components are removed to evaluate their effectiveness. We use the solved rate of problems as the metric and the base model is LLaMA-3.1-8B-Instruct. }
\label{tab:performance-breakdown}
\small
\setlength{\tabcolsep}{4pt} 

\begin{tabular}{l|cc|cc}
\toprule
\hline
 & \multicolumn{2}{c|}{GSM8K} & \multicolumn{2}{c}{AIME24} \\
\hline
Zero-shot CoT                                   & \multicolumn{2}{c|}{76.6}& \multicolumn{2}{c}{6.7}\\
\hline
 Rollouts &  8& 16& 8& 16\\
 \hline
Iterative Self-Refine& 78.9 & 78.2 & 6.7  & 6.7  \\
SR-MCTS without PPRM& 82.0& 81.4& 6.7& 6.7\\
\textbf{SR-MCTS + PPRM~(Ours)}      & \textbf{95.0}& \textbf{96.1}& \textbf{16.7}& \textbf{26.7}\\
\hline
\bottomrule
\end{tabular}
\end{table}

Notably, on the more challenging AIME2024 dataset, both iterative Self-Refine and SR-MCTS without PPRM demonstrate limited search efficiency. However, SR-MCTS with PPRM can significantly improve the solved rate of problems from 6.7\% (2/30) to 16.7\% (5/30) and 26.7\% (8/30) with rollouts of 8 and 16, respectively. The results underscore the efficacy of combining the Self-Refine method with PPRM when addressing complex problems. The contrast between self-reward and PPRM underscores the importance of designing reward mechanisms that can more effectively guide the search process. PPRM provides a more holistic incentive to the model, thus fostering more effective problem-solving strategies.

 



\subsection{Scaling Study}
\begin{figure*}[!tbp]
    \label{test-time-scaling-2}
    \centering
    \includegraphics[width=0.87\linewidth]{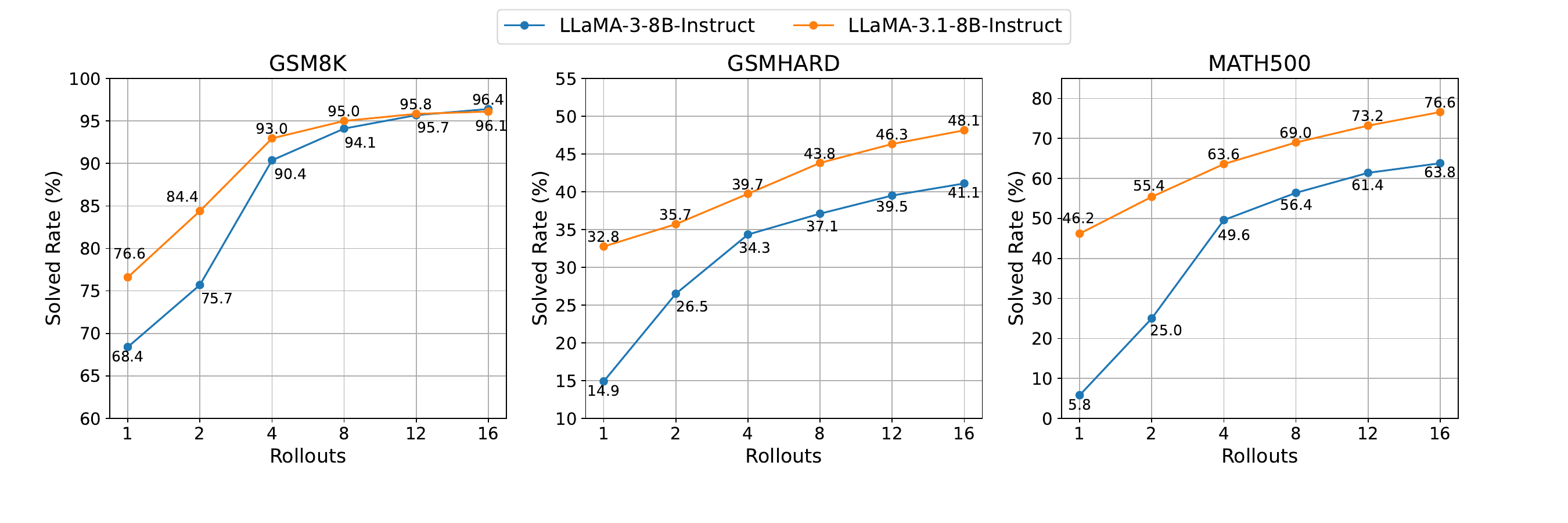}
    \caption{Scaling of inference-time rollouts}
    \label{fig:test-time-scaling}
\end{figure*}

To explore the potentials and trends of the scaling with rollouts in inference-time, we depict the solved rate of problems with rollouts in three benchmarks with different difficulty levels. Analyzing the performance alongside Figure~\ref{fig:test-time-scaling}, the increment in the number of rollouts consistently enhances model performance across various benchmarks, and the extent of these improvements differs depending on the benchmark's complexity and the base model's reasoning capability.  These curves underscore that the performance of the \texttt{LLaMA-Berry} framework benefits from scaling up rollouts during inference, similar to the observations in~\citet{openAI-o1}. However, there are ceiling limitations, as seen in the GSM8K dataset, which suggest that the base model's capabilities in both reasoning and refinement play a crucial role in determining the overall performance.

\section{Related Works}

\noindent\textbf{Reward models for reasoning. }Reliable reward models~\citep{MindStar, MathShepherd, GLoRe, lightman2023let, reward-step-by-step} can effectively distinguish desirable responses from undesirable ones, which is especially important in complex reasoning. The outcome reward model (ORM) are trained with the final results of the reasoning paths. As the rewards are determined by the final answers, ORM may suffer from coarse supervision and misalignment issues. In contrast, process reward model (PRM) provides step-wise reward signals that are easier to interpret and guide the models to follow the CoTs. Therefore, PRM is generally considered to be more effective~\citep{lightman2023let}. However, the success of PRM relies on extensive manually annotated data~\citep{luo2024improvemathematicalreasoninglanguage, GLoRe}, which is time-consuming and still faces the challenge of the volatility of absolute reward scores. Pairwise Preference Reward Model (PPRM) in ~\texttt{LLaMA-Berry} converts absolute scoring into preference prediction task, which then brings robust reward signals via EBC method.

\noindent\textbf{Tree search reasoning. }Sampling diverse reasoning paths~\citep{brown2024large} has demonstrated its effectiveness in enhancing the probability of finding the correct answers. Self-Consistency~\citep{wang2022self} samples a complete path each time while tree search methods like Tree-of-Thought (ToT)~\citep{ToT} and Monte Carlo Tree Search (MCTS)~\citep{chen2024alphamath, chen2024step, luo2024improve, Alphazero-like, IterativePreferenceLearning, No-Train-Still-Gain, Don't-throw-away-your-value-model, Imagination-Searching-and-Criticizing, Everything-of-Thoughts} extend multiple steps to optimize step answers and ultimately obtain the optimal solution. Additionally, Self-Refine~\citep{Self-Refine-first} method has become a recent focus. Self-verification~\citep{gero2023self, weng2022large} and rStar~\citep{Mutual-Reasoning} utilize the inherent capabilities of the model to iteratively explore and refine answers. However, the performance of Self-Refine is typically constrained by the inherent capabilities of the model, especially for small language models (SLMs) with significantly weaker Self-Refine abilities~\citep{NEURIPS2023_91edff07}. \citet{DoT} suggests that the mathematical reasoning abilities of LLMs can be enhanced by treating the refinement process as a directed acyclic graph (DAG) through multi-agent collaboration. In our approach, we combine MCTS with Self-Refine to explore potential solutions and a global win-loss matrix is then constructed in the form of a directed graph to calculate the final quantile scores.

\section{Conclusion}

This research addresses challenges in complex mathematical reasoning, particularly at the Olympiad level, by enhancing the accuracy and efficiency of the search for reasoning paths. By introducing Self-Refine applied to Monte Carlo Tree Search (SR-MCTS), the \texttt{LLaMA-Berry} framework significantly improves the efficiency of solution generation by LLMs. Additionally, the Pairwise Preference Reward Model (PPRM) constructs preferences between solutions rather than merely scoring outcomes, calculating the final global quantile score using the enhanced Borda Count (EBC) method. Evaluation results demonstrate that \texttt{LLaMA-Berry} outperforms baseline approaches on benchmarks like GSM8K and MATH, and achieves competitive performance compared to closed-source models on Olympiad-level benchmarks such as AIME2024.

\section*{Limitation}

The \texttt{LLaMA-Berry} framework has demonstrated strong performance in reasoning tasks, but there are still some challenges in practical applications. First, methods such as Monte Carlo Tree Search (MCTS) and Self-Refine have high computational costs. These techniques demand significant computational resources, which may limit the feasibility of deployment in environments with constrained computational capacity. As for summarizer of solutions, rule-based heuristics methods like self-consistency, major voting and mutual reasoning have shown a constraint to the ceiling search performance of MCTS. Thus, we aim to develop a learning-based summarizer as ~\citet{openAI-o1} does to further enhance the search efficiency.

Furthermore, the current evaluation of the \texttt{LLaMA-Berry} framework has primarily focused on mathematical reasoning benchmarks, resulting in a relatively narrow assessment scope. As a result, its performance in broader domains, such as general knowledge, symbolic logic tasks, and multimodal applications, has not been sufficiently validated. In future work, we aim to improve the framework by evaluating it on a more diverse set of tasks to enhance its applicability.

Lastly, most experiments so far have utilized relatively small open-source models, with limited testing on larger or closed-source models. In future research, we plan to investigate the performance of \texttt{LLaMA-Berry} on larger models, particularly addressing challenges related to scaling and performance optimization.



\bibliography{main}

\clearpage
\appendix
\setcounter{table}{0} 
\renewcommand{\thetable}{A\arabic{table}}
\setcounter{figure}{0}
\renewcommand{\thefigure}{A\arabic{figure}}
\setcounter{equation}{0}
\renewcommand{\theequation}{A\arabic{equation}}
\newpage
\newpage
\appendix

\section{Hyper-paramerer settings}
\label{sec:Hyper-paramerer settings}

All hyper-parameter settings for Section~\ref{sec:Evaluation} are listed in Table~\ref{tab:hyperparameter}.
\begin{table}[htbp!]
\caption{Hyper-parameter settings}
\centering
\resizebox{1\columnwidth}{!}{\begin{tabular}{c|c|c}
\toprule
\hline
\textbf{Hyper-parameter} & \textbf{Description} & \textbf{Value} \\ \hline
$\alpha$ & Balance coefficient between local and global & 0.9 \\ \hline
$c$ & Exploration coefficient in UCT function & 1.4 \\ \hline
$\gamma$ & Discount factor for backpropagation & 0.1 \\ \hline
$max\_child$ & Dynamic pruning hyper-parameter in search & 2 \\ \hline \bottomrule
\end{tabular}}
\label{tab:hyperparameter}
\end{table}
\section{Details of Grading and Metrics}
\label{sec:grading and metrics}

\textbf{Grading. } We follow the correctness evaluation method proposed in PRM800K~\citep{lightman2023let} to score the answers generated by the algorithm. For the mathematical solutions proposed by the algorithm and their corresponding ground truth, we inform the model of the expected response format in a prompt. The answer's formula or value is extracted by matching the response format with predefined rules. If the model fails to follow the expected format in the prompt and the rule-based extraction fails, the solution is directly judged as inconsistent with ground truth.

For the extracted label, we score the answer based on the following criteria. The answer is considered consistent with the ground truth label and passes the evaluation if at least one of the criteria is met:

\begin{enumerate}
    \item The answer label string is exactly equal to the ground truth label string in terms of literal value.
    \item Both the answer label and the ground truth label can be converted to floating-point numbers, and the difference between the two values is less than \(1 \times 10^{-6}\).
    \item Following the criterion proposed in PRM800K~\citep{lightman2023let}, we use the Sympy library~\citep{meurer_sympy_2017} to simplify the difference between the expression corresponding to the answer label, denoted as \(a\), and the expression corresponding to the ground truth label, denoted as \(b\). If the simplification yields \(a - b \iff 0\), the criterion is satisfied.
\end{enumerate}




\noindent\textbf{Metrics. } To provide a comprehensive and robust evaluation metric, we adopt \textbf{major@k} and \textbf{rm@k} as the evaluation metrics.

\begin{itemize}
    \item \textbf{rm@k} represents reward model best-of-N among $k$ sampled response~\citep{yang2024qwen2}.

    \item \textbf{major@k}~\citep{kuncheva2014combining}, also abbreviated as \textbf{maj@k}, is defined as the fraction of tasks where a majority of the top \(k\) samples generated return the same correct solution. This metric focuses on consistency across multiple generated answers. The majority weight is calculated on the solution's extracted labels.
    \item The  \textbf{solved rate of problems}~\citep{lightman2023let,luo2024improvemathematicalreasoninglanguage} refers to the percentage of problems who have solutions meet the evaluation criteria.
\end{itemize}
For results without a subscript, the score represents the greedy evaluation using the default prompt of the base model. Results with a notation indicate the use of the corresponding prompt engineering technique, and those marked with \textbf{self-consistency} use the self-consistency aggregation method.

For closed-source models, we report the scores from existing official technical reports~\citep{anthropic2024claude3, reid2024gemini, OpenAIo12024} or dataset-provided results, with no modifications.
\section{Details of PPRM Trainging}
\label{sec:Details of PPRM Trainging}

\subsection{Overview}
The design goal of the PPRM is to integrate the properties of both PRM and ORM, providing a more nuanced preference prediction between any two solution answers. We attempt to utilize reinforcement learning methods for training, leveraging the model's instruction-following capability to predict the relative merits of pairs of problem-solving answers. This will further enable the use of the EBC method to evaluate the global quantile scores of mathematical solutions.

\subsection{Data Collection}

\begin{figure*}[t]
    \centering
\includegraphics[width=0.85\linewidth]{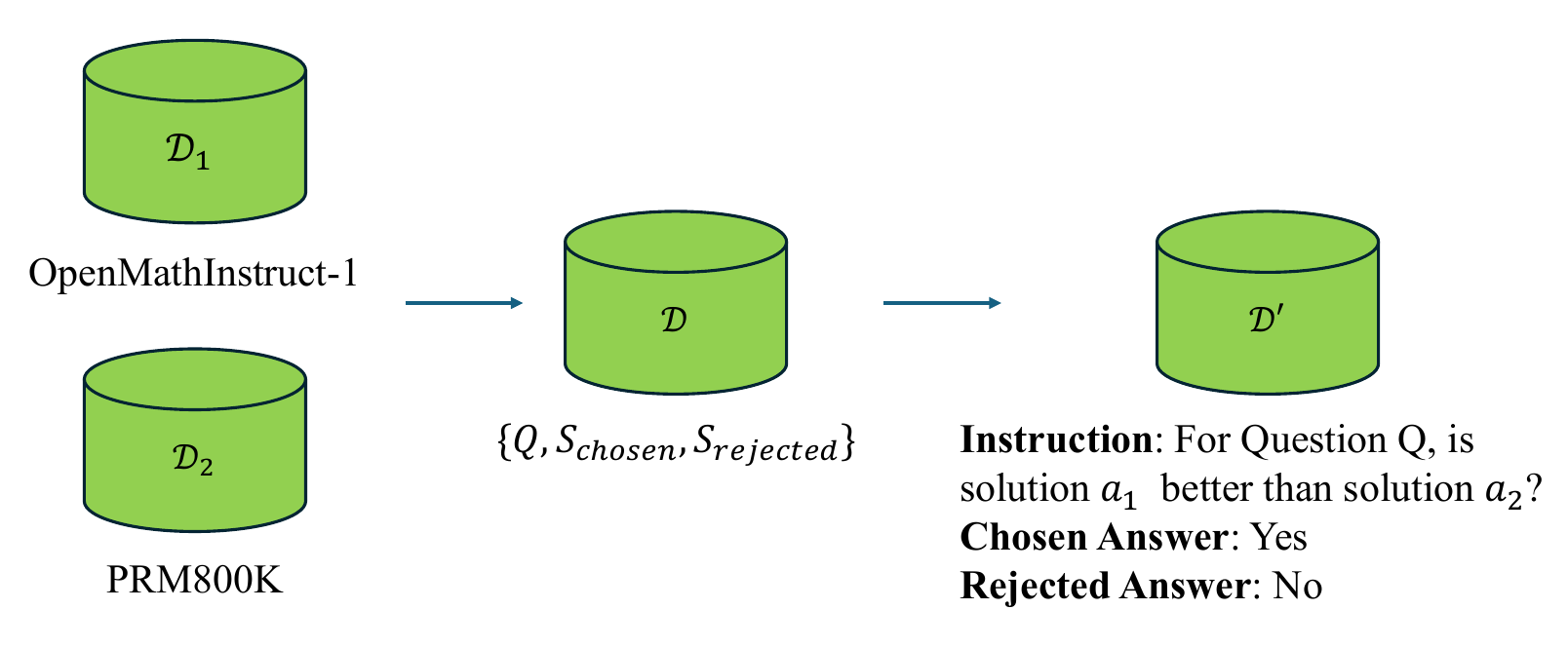}
    \caption{Dataset Construction of PPRM.}
    \label{fig:pprm_instruction}
\end{figure*}

Our data synthesis is derived from two datasets: PRM800K~\citep{lightman2023let} and OpenMathInstruct-1~\citep{OpenMathInstruct-1}. The PRM800K dataset, collected from the MATH dataset, comprises a substantial number of step-divided problem-solving answers, with manual quality annotations for each step. We primarily utilize this dataset to generate pairs of answers for comparative analysis based on step-wise process quality. The OpenMathInstruct-1 dataset incorporates data from the GSM8K and MATH datasets, which have been manually annotated for outcome correctness. We use this dataset to synthesize pairs for comparative analysis based on outcome quality.

In processing PRM800K, for a given problem, we first sample steps of varying quality annotations from the step-wise dataset to construct a complete reasoning path. For pairs with the same final step annotation, paths composed of higher-quality steps are considered superior to those with lower-quality steps. In cases where the final step annotations are not identical, the reasoning path with the superior final step annotation is regarded as better.

During the processing of OpenMathInstruct-1, we exclusively utilize samples without Python interpreter calls. For the same problem, we sample pairs composed of outcomes with higher and lower quality annotations.

Notably, we filtered out any data samples from PRM800K and OpenMathInstruct-1 that may overlap with the GSM8K and MATH test sets, especially the PRM800K. Ultimately, we formed a dataset of 7,780,951 entries for training the PPRM model.

\subsection{Direct Preference Pair Construction}

For all pairs, we frame the inquiry as "For Question \(Q\), is solution \(a_1\) better than solution \(a_2\)?" If solution \(a_1\) is deemed superior to solution \(a_2\), we label it as 'Yes'; otherwise, it is labeled as 'No'.

In this manner, we transform the ordinal relationship prediction task into a question-answer format, employing the Direct Preference Optimization (DPO) method for model training through reinforcement learning from human feedback (RLHF). This approach aims to enhance the model's capability to follow instructions in predicting the relative merits of pairs of problem-solving answers.

\subsection{RLHF Training}
\begin{figure*}[t]
    \centering
    \includegraphics[width=1\linewidth]{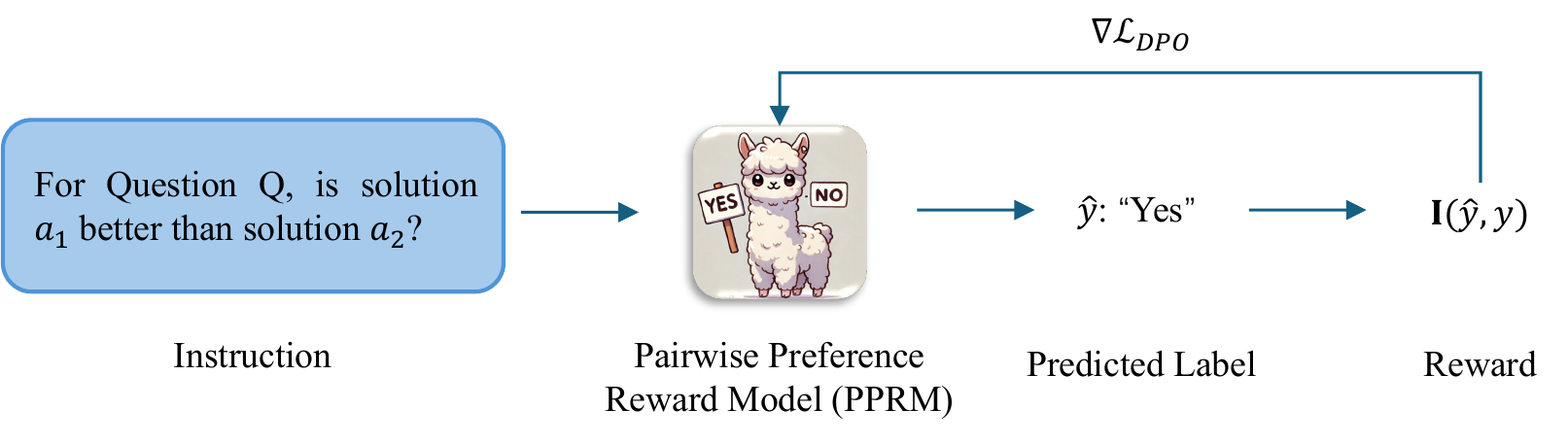}
    \caption{RLHF training of PPRM.}
    \label{fig:pprm_training}
\end{figure*}

We apply the DPO method to train the Gemma2-2B-it model using RLHF which is shown in Figure \ref{fig:pprm_training}. The loss function for DPO is structured as $\mathcal{L}_{\mathrm{DPO}}(\pi_\theta;\pi_{\mathrm{ref}})=-\mathbb{E}\left[\log\sigma\left(\beta\log\frac{\pi_\theta(y_w\mid x)}{\pi_{\mathrm{ref}}(y_w\mid x)} - \beta\log\frac{\pi_\theta(y_l\mid x)}{\pi_{\mathrm{ref}}(y_l\mid x)}\right)\right]$, where $\sigma$ is the logistic function, $\beta$ is a parameter controlling the deviation from the base reference policy $\pi_{ref}$, namely the initial model $\pi^{LLM}$. In practice, the language model policy $\pi_{\theta}$ is also initialized to $\pi^{LLM}$. For one answer, denoted as $y_w \succ y_l | x$ where $y_w$ and $y_l$ denote the preferred and dispreferred completion amongst $(\hat{y}, y)$ respectively.













\section{Details of Berry-Tree Inference Framework}

\subsection{Overview}
\texttt{Berry-Tree} is an inference framework specifically designed for complex multi-model reasoning tasks, addressing the need to improve inference efficiency and accuracy in intricate tree search processes of LLM's mathematical reasoning. This framework is particularly suited for large-scale reasoning tasks that involve the integration of multiple models and algorithms. By incorporating advanced tree search methods especially Monte Carlo Tree Search~(MCTS), robust concurrency handling, and high-performance inference engines, \texttt{Berry-Tree} significantly enhances inference speed and resource utilization of LLM's mathematical reasoning process. This section provides a explanation of the system architecture and key technologies of \texttt{Berry-Tree}, along with a preliminary performance evaluation results.

\subsection{System Architecture Overview}

\begin{figure*}
    \centering
    \includegraphics[width=0.8\linewidth]{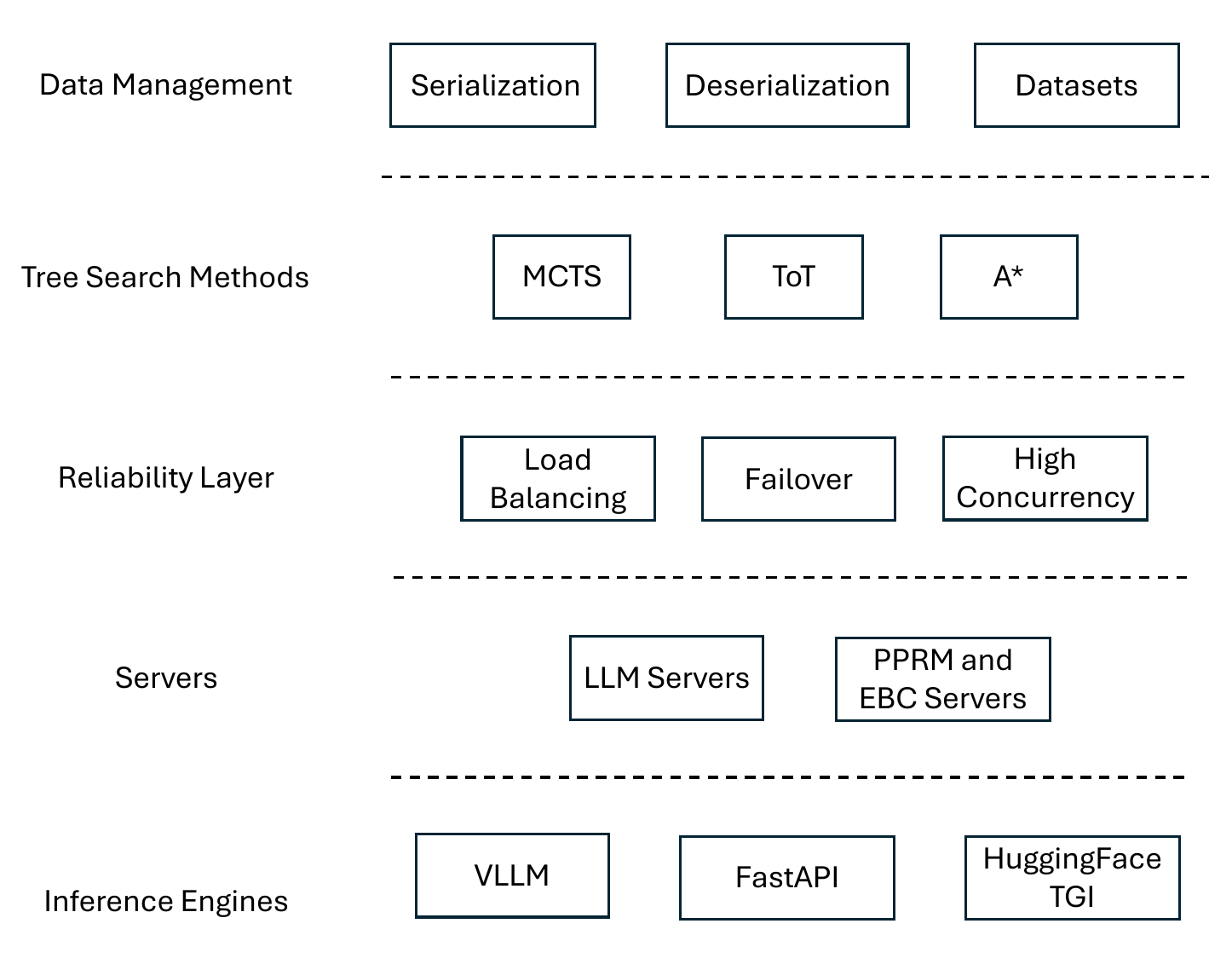}
    \caption{Architecture design of \texttt{Berry-Tree}}
    \label{fig:berry}
\end{figure*}

Figure~\ref{fig:berry} demonstrates the architecture of \texttt{Berry-Tree} which is divided into several layers, each handling different functional requirements. The \textbf{Data Management Layer} is responsible for the serialization and deserialization of data, ensuring efficient data read/write operations across models and systems and the ablity of recovering the search process from serialized data. The \textbf{Tree Search Methods Layer} incorporates MCTS (Monte Carlo Tree Search), ToT (Tree of Thoughts), and A* algorithms to optimize the inference process and explore multiple reasoning paths. Additionally, \texttt{Berry-Tree} includes a \textbf{Reliability Layer}, which ensures load balancing and failover support in highly concurrent scenarios, guaranteeing the stability of inference tasks. Finally, the \textbf{Inference Engines Layer} integrates efficient inference engines such as vLLM, FastAPI, and HuggingFace TGI to enable parallelized and efficient task processing.

\subsection{Key Technical Components}

\textbf{Data Management.} \texttt{Berry-Tree} employs serialization and deserialization techniques, specifically using formats including JSON and CSV, to efficiently store, transfer and recover checkpoint data from tree search reasoning processes. The framework stores this data along with hash values to ensure integrity and allows for quick restoration of tree search states in memory when needed. Furthermore, \texttt{Berry-Tree} leverages the HuggingFace Datasets library to handle core dataset inputs for both training and inference. It supports seamless loading of benchmark datasets such as GSM8K and MATH from the HuggingFace Hub, enhancing its flexibility and ensuring compatibility with diverse reasoning tasks.

\noindent\textbf{Tree Search Methods.} \texttt{Berry-Tree} support multiple tree search algorithms, with MCTS being central to handling large-scale complex reasoning tasks by leveraging random simulation and statistical analysis to optimize the search space. The Tree of Thoughts (ToT) extends the exploration depth and breadth of reasoning paths, helping the system manage uncertainty. And A* can offer a efficient heuristic search capabilities.

\noindent\textbf{Reliability Design.} To ensure the stability and continuity of inference tasks, \texttt{Berry-Tree} incorporates load balancing and failover mechanisms. During high-concurrency operations, the load balance componet distributes workloads across different inference servers, preventing server overload. The failover mechanism ensures that tasks can seamlessly recover and transition to backup servers in case of partial server failures.

\noindent\textbf{Server Architecture.} The framework's server architecture is divided into two segments: one dedicated to executing Large Language Model (LLM) inference and the other designated explicitly for handling PPRM (Pairwise Preference Reward Model) with EBC (Enhanced Borda Count) method. This modular design allows the framework to allocate computational resources flexibly, improving overall efficiency.

\noindent\textbf{Inference Engines.} Inference engines of \texttt{Berry-Tree} include VLLM , HuggingFace Transformers warpped by FastAPI, and HuggingFace TGI. These engines collectively enable the system to maintain high efficiency and stability while handling multi-model inference tasks, with robust support for high-concurrency demands.

\subsection{Preliminary Performance Evaluation}

In a preliminary performance evaluation, we utilize 16 A100 GPUs to run the LLaMA3.1-8B-instruct model for large-scale inference tasks, while 4 A100 GPUs are used to run the Gemma2-2B-it model as PPRM servers. The test dataset consists of 1319 GSM8K test samples.

We conduct 16 rollouts to parallelize the inference tasks of \texttt{LLaMA-Berry} via \texttt{Berry-Tree}. The results indicate that the total inference time is 1 hour and 25 minutes, with an average inference time of approximately 3.87 seconds per sample. These results demonstrate a strong parallel inference capabilities of \texttt{Berry-Tree} under the given hardware configuration.



\section{Scaling Study on Inference-time Token Overheads}

\begin{figure*}[t]
    \centering
    \includegraphics[width=0.9\linewidth]{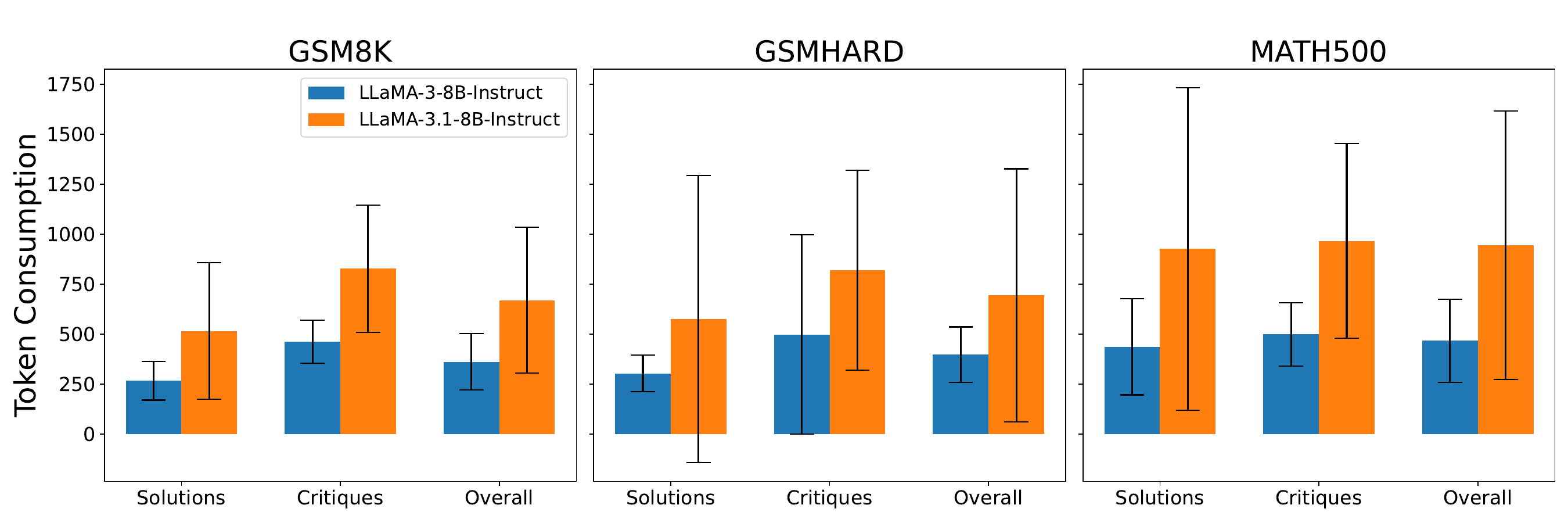}
    \caption{Average token consumption comparison across datasets, error bar stands for standard deviation.}
    \label{fig:test-time-token-scaling}
\end{figure*}



\begin{figure*}[!]
    \centering
    \includegraphics[width=0.9\linewidth]{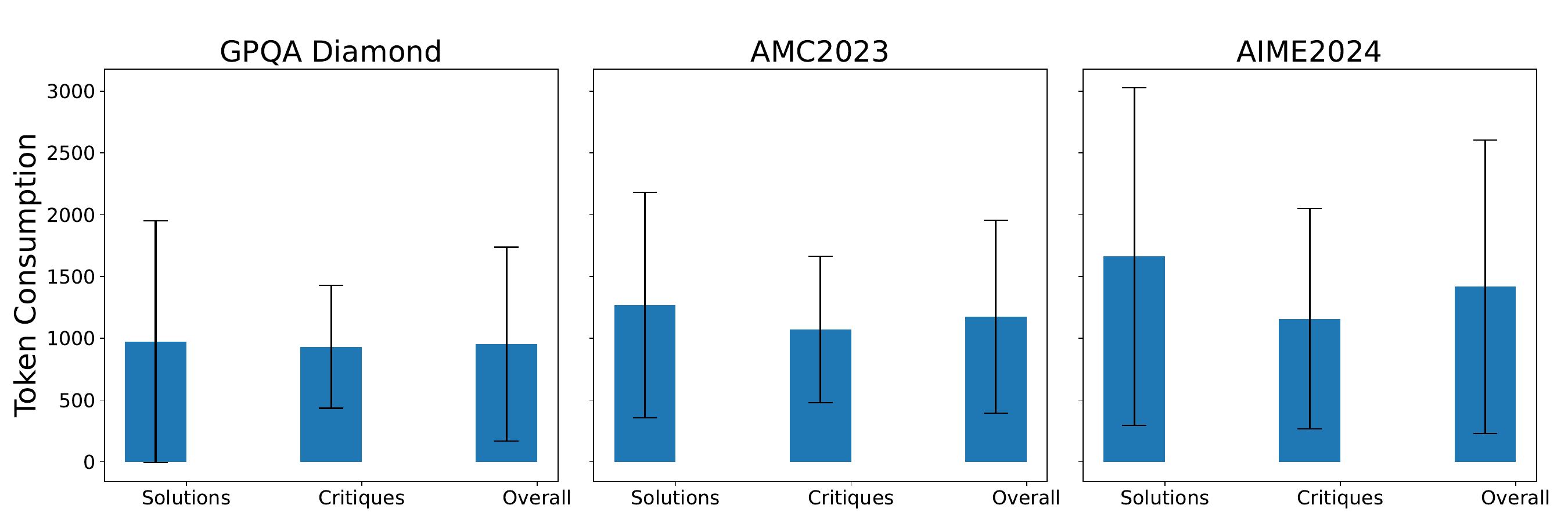}
    \caption{Average token consumption for LLaMA-3.1-8B-Instruct across olympiad-level datasets, error bar stands for standard deviation.}
    \label{fig:test-time-olympiad-token-scaling}
\end{figure*}

Comparing LLaMA-3-8B-Instruct and LLaMA-3.1-8B-Instruct across GSM8K, GSMHARD, and MATH500, LLaMA-3.1-8B-Instruct consistently consumes more tokens across all categories—Solutions, Critiques, and Overall. This Figure~\ref{fig:test-time-token-scaling} result reflects LLaMA-3.1-8B-Instruct's tendency to generate more detailed and comprehensive outputs with increased overhead. While this likely improves solution quality, it also introduces greater resource demands and variability in token usage, highlighting a trade-off between accuracy and computational resources. LLaMA-3.1-8B-Instruct is thus better suited for tasks prioritizing precision over speed. As shown in Figure ~\ref{fig:test-time-olympiad-token-scaling}, Token overheads during inference also scale with task difficulty across different olympiad-level benchmarks with LLaMA-3.1-8B-Instruct. AIME2024 exhibits the highest token consumption with significant variability, reflecting the complexity of its solution paths. In contrast, GPQA Diamond shows lower overall overhead, while AMC2023 falls in between, with moderate token consumption and less variability than AIME2024 but still notable.
\section{Future Work}
\label{sec:limit_future_work}


\texttt{LLaMA-Berry} holds significant potential for further development. First, we plan to enhance its multimodal capabilities, enabling it to better handle complex problems like VQA and mathematic geometric problems that require visual, auditory, or even tactile perception. For instance, in tasks involving visual reasoning, such as geometric problems, \texttt{LLaMA-Berry} could be extended to integrate image recognition and analysis capabilities, thereby assisting in solving challenges related to spatial relationships and shape recognition. Furthermore, we aim to generalize its application across other scientific domains, improving its performance in disciplines such as physics, chemistry, and biology. In physics, for example, it could be utilized to address complex micro- and macroscopic dynamics problems; in chemistry, it may assist in molecular structure prediction and drug design; and in biology, it could potentially be used for genome analysis and disease prediction. Moreover, \texttt{LLaMA-Berry} can also offer technical support for other interdisciplinary fields, such as meteorology, environmental science, and materials science, by integrating multimodal data to enhance predictive accuracy. At the same time, we will focus on how \texttt{LLaMA-Berry} can further enhance AI safety. For instance, \texttt{LLaMA-Berry} could be leveraged to design more robust safety and risk assessment mechanisms. Besides, the generation of responses could be guided by a safety-trained PPRM to produce outputs with higher safety standards. These advancements not only pave the way for broader scientific applications of \texttt{LLaMA-Berry} but also offer new possibilities for enhancing AI safety and promoting its widespread adoption.

\section{Case Study}

\begin{figure*}[!t]
    \centering
    \includegraphics[width=0.66\linewidth]{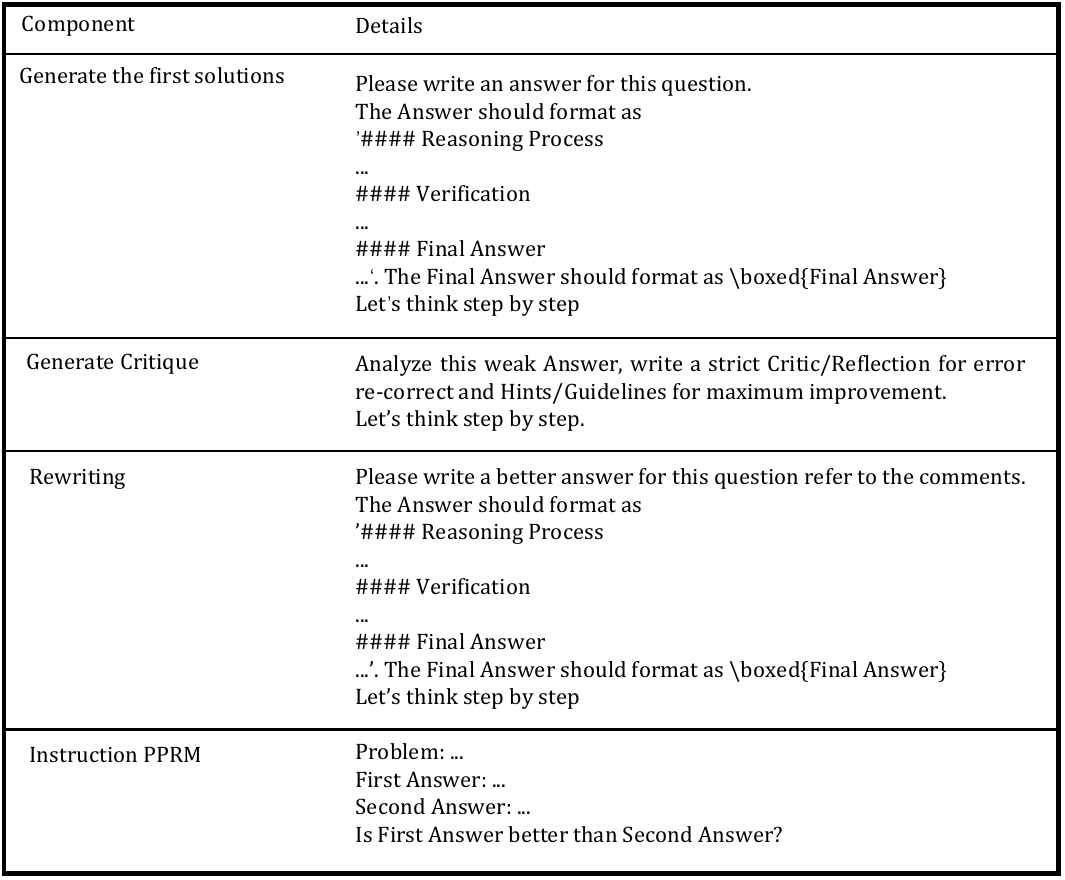}
    \caption{Prompts for LLaMA-3.1-8B-Instruct}
    \label{fig:prompts-for-llama}
\end{figure*}

\begin{figure*}[!t]
    \centering
    \includegraphics[width=0.66\linewidth]{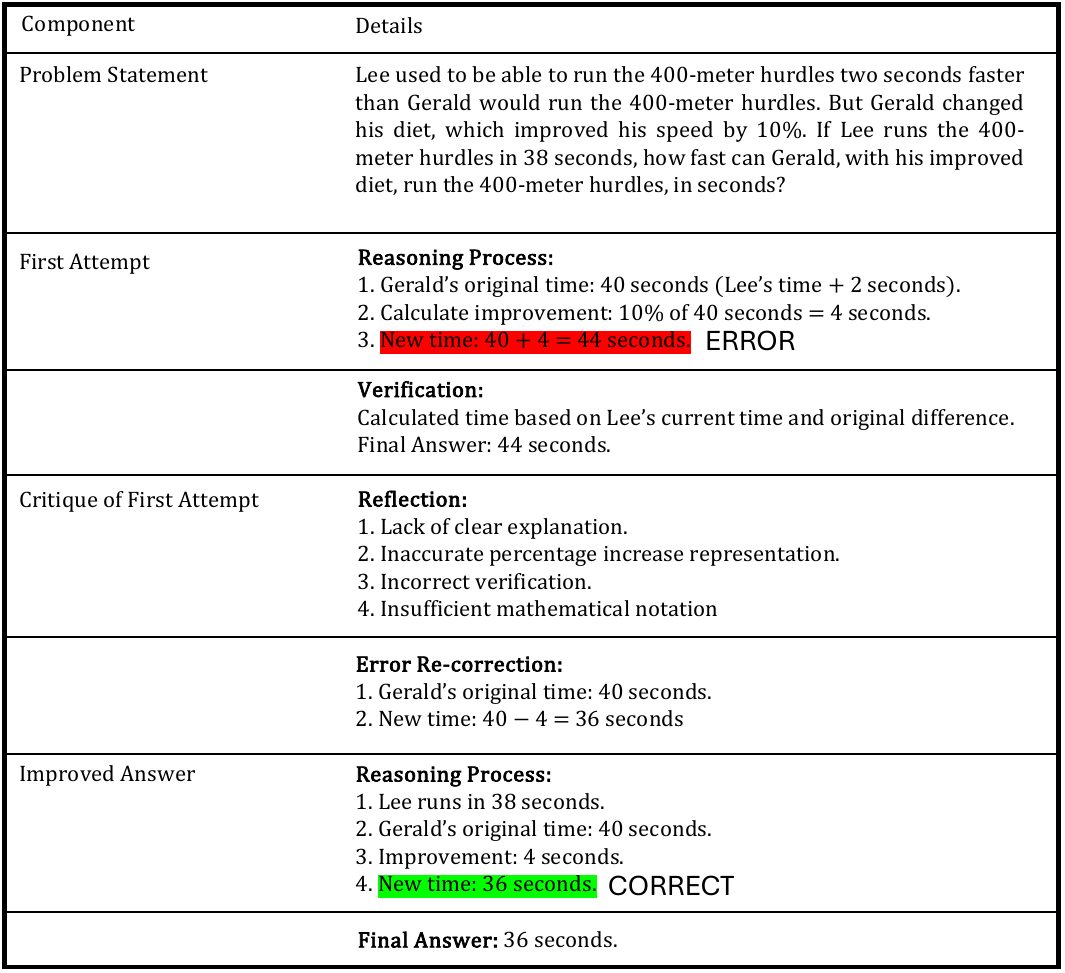}
    \caption{Problem-solving example}
    \label{fig:solving-example}
\end{figure*}

The prompts utilized in \texttt{LLaMA-Berry} for the LLaMA-3.1-8B-Instruct model are presented in Figure~\ref{fig:prompts-for-llama}. Additionally, Figure~\ref{fig:solving-example} provides a detailed breakdown of problem-solving examples derived from the GSM8K dataset.

 \section{Convergence Analysis of the Enhanced Borda Count (EBC) Method}
\label{sec:Proof of Convergence of the Enhanced Borda Count (EBC) Method}
In this appendix, we present a formal discussion on how the quantile scores, as evaluated by the Enhanced Borda Count (EBC) method, converge to the true quantile scores of solutions within the actual quality distribution as the number of samples increases.

\subsection{Definitions and Assumptions}

\textbf{Solution Set.} Let $A = \{ a_1, a_2, \ldots, a_n \}$ be a finite set of $n$ solutions (answers).
    
\noindent\textbf{True Quantile Scores.} Each solution $a_i$ has a true quality score $Q^*(a_i) \in \mathbb{R}$, drawn from a continuous distribution $P_Q$.
    
\noindent\textbf{True Ranking.} The true ranking $R^*$ is induced by ordering the solutions in decreasing order of their true quantile scores $Q^*(a_i)$.

\noindent\textbf{True Pairwise Preference Probabilities.} The true probability that solution $a_i$ is preferred over $a_j$ is defined as:
    \begin{equation}
    \begin{aligned}
    &P^*(a_i \succ a_j) = \\&\mathbb{P}(Q^*(a_i) > Q^*(a_j)) + \tfrac{1}{2} \mathbb{P}(Q^*(a_i) = Q^*(a_j)).
    \end{aligned}
    \end{equation}
    Given the continuity of $P_Q$, we have $\mathbb{P}(Q^*(a_i) = Q^*(a_j)) = 0$, so:
    \begin{equation}
    P^*(a_i \succ a_j) =
    \begin{cases}
    1, & \text{if } Q^*(a_i) > Q^*(a_j), \\
    0, & \text{if } Q^*(a_i) < Q^*(a_j).
    \end{cases}
    \end{equation}
    
\noindent\textbf{Estimated Preference Probabilities.} For $T$ independent samples, the estimated preference probability is:
    \begin{equation}
    P_T(a_i \succ a_j) = \frac{1}{T} \sum_{t=1}^T X_t^{(i,j)},
    \end{equation}
    where $X_t^{(i,j)}$ are independent Bernoulli random variables with success probability $P^*(a_i \succ a_j)$.
    
\noindent\textbf{Convergence Assumption.} As $T \to \infty$, $P_T(a_i \succ a_j)$ converges almost surely to $P^*(a_i \succ a_j)$.
    
\noindent\textbf{Preference Matrix.} Construct the estimated preference matrix $M_T$ and the true preference matrix $M^*$ as:
    \begin{equation}
    \begin{aligned}
    &M_T[i, j] = 
    \begin{cases}
    1, & \text{if } P_T(a_i \succ a_j) \geq 0.5, \\
    0, & \text{otherwise.}
    \end{cases}\\
    &M^*[i, j] = 
    \begin{cases}
    1, & \text{if } P^*(a_i \succ a_j) = 1, \\
    0, & \text{if } P^*(a_i \succ a_j) = 0.
    \end{cases}
    \end{aligned}
    \end{equation}
    
\noindent\textbf{Transitive Closure.} Let $C_T$ and $C^*$ be the transitive closures of $M_T$ and $M^*$, respectively.
    
\noindent\textbf{Borda Counts.} Compute the Borda count for each solution:
    \begin{equation}
    B_T(a_i) = \sum_{j \neq i} C_T[i, j], \quad B^*(a_i) = \sum_{j \neq i} C^*[i, j].
    \end{equation}
    
\noindent\textbf{Estimated Ranking and Quantile Scores.}
    \begin{itemize}
        \item Estimated ranking $R_T$: Order solutions by decreasing $B_T(a_i)$.
        \item Estimated quantile score:
        \begin{equation}
        Q_g(a_i) = 1 - \frac{\text{rank}_T(a_i) - 1}{n - 1}.
        \end{equation}
        \item True quantile score:
        \begin{equation}
        Q^*_g(a_i) = 1 - \frac{\text{rank}^*(a_i) - 1}{n - 1}.
        \end{equation}
    \end{itemize}

\subsection{Objective}

To prove that as $T \to \infty$, the estimated quantile scores $Q_g(a_i)$ converge almost surely to the true quantile scores $Q^*_g(a_i)$:
\begin{equation}
\lim_{T \to \infty} Q_g(a_i) = Q^*_g(a_i), \quad \forall a_i \in A.
\end{equation}

\subsection{Proof}

\textbf{Convergence of Estimated Preference Probabilities.}

By the Strong Law of Large Numbers (SLLN), since $X_t^{(i,j)}$ are i.i.d. Bernoulli random variables with success probability $P^*(a_i \succ a_j)$, we have:
\begin{equation}
\lim_{T \to \infty} P_T(a_i \succ a_j) = P^*(a_i \succ a_j) \quad \text{almost surely}.
\end{equation}

\noindent\textbf{Convergence of Preference Matrix $M_T$ to $M^*$.}

Since $P_T(a_i \succ a_j)$ converges to $P^*(a_i \succ a_j)$ and $P^*(a_i \succ a_j) \in \{0, 1\}$, for sufficiently large $T$, we have:
\begin{equation}
M_T[i, j] = M^*[i, j], \quad \forall i \neq j,
\end{equation}
almost surely.

\textit{Justification}: Because $P_T(a_i \succ a_j)$ converges to either 0 or 1, and $P_T(a_i \succ a_j) \neq 0.5$ almost surely for large $T$.

\noindent\textbf{Convergence of Transitive Closure $C_T$ to $C^*$.}

The transitive closure is a deterministic function of the preference matrix. Therefore, since $M_T \to M^*$, it follows that:
\begin{equation}
C_T \to C^* \quad \text{as } T \to \infty,
\end{equation}
almost surely.

\noindent\textbf{Convergence of Borda Counts $B_T(a_i)$ to $B^*(a_i)$.}

Given that $C_T \to C^*$, the Borda counts converge:
\begin{equation}
B_T(a_i) = \sum_{j \neq i} C_T[i, j] \to B^*(a_i) = \sum_{j \neq i} C^*[i, j],
\end{equation}
almost surely.

\noindent\textbf{Convergence of Estimated Ranking $R_T$ to True Ranking $R^*$.}

Since the Borda counts $B_T(a_i)$ converge to $B^*(a_i)$, and assuming that all $Q^*(a_i)$ are distinct (due to the continuity of $P_Q$), the rankings induced by $B_T(a_i)$ converge to the true rankings:
\begin{equation}
R_T \to R^* \quad \text{as } T \to \infty,
\end{equation}
almost surely.

\noindent\textbf{Convergence of Quantile Scores $Q_g(a_i)$ to $Q^*_g(a_i)$.}

Since $\text{rank}_T(a_i) \to \text{rank}^*(a_i)$, we have:
\begin{equation}
\begin{aligned}
&Q_g(a_i) = 1 - \frac{\text{rank}_T(a_i) - 1}{n - 1} \to \\&Q^*_g(a_i) = 1 - \frac{\text{rank}^*(a_i) - 1}{n - 1}
\end{aligned}
\end{equation}
almost surely.

\noindent\textbf{Conclusion.}

Therefore, we have shown that:
\begin{equation}
\lim_{T \to \infty} Q_g(a_i) = Q^*_g(a_i), \quad \forall a_i \in A,
\end{equation}
which means that the EBC method's estimated quantile scores converge almost surely to the true quantile scores of the solutions.

\subsection{Finite Sample Analysis and Convergence Rate}

\textbf{Lemma 1: Hoeffding's Inequality for Preference Probability Estimates.}

For each pair $(a_i, a_j)$, $P_T(a_i \succ a_j)$ is the sample mean of $T$ i.i.d. Bernoulli trials with success probability $P^*(a_i \succ a_j)$. By Hoeffding's inequality:
\begin{equation}
\begin{aligned}
\mathbb{P}\left( \left| P_T(a_i \succ a_j) - P^*(a_i \succ a_j) \right| \geq \epsilon \right) \\\leq 2 \exp(-2 T \epsilon^2).
\end{aligned}
\end{equation}

\noindent\textbf{Lemma 2: Uniform Convergence over All Pairs.}

Apply the union bound over all $N = n(n - 1)/2$ pairs:
\begin{equation}
\begin{aligned}
\mathbb{P}\left( \exists (i, j): \left| P_T(a_i \succ a_j) - P^*(a_i \succ a_j) \right| \geq \epsilon \right)\\ \leq N \cdot 2 \exp(-2 T \epsilon^2).
\end{aligned}
\end{equation}
Set the right-hand side equal to $\delta$ to solve for $T$:
\begin{equation}
T \geq \frac{1}{2 \epsilon^2} \ln\left( \frac{2 N}{\delta} \right).
\end{equation}

\noindent\textbf{Lemma 3: Correctness of Preference Matrix with High Probability.}

Given that $P^*(a_i \succ a_j) \in \{0, 1\}$, for any $0 < \epsilon < 0.5$, if:
\begin{equation}
\left| P_T(a_i \succ a_j) - P^*(a_i \succ a_j) \right| < \epsilon,
\end{equation}
then $M_T[i, j] = M^*[i, j]$ because $P_T(a_i \succ a_j)$ will be greater than $0.5$ when $P^*(a_i \succ a_j) = 1$, and less than $0.5$ when $P^*(a_i \succ a_j) = 0$.

\noindent\textbf{Lemma 4: Probability of Correct Ranking.}

From Lemma 2 and 3, with probability at least $1 - \delta$, $M_T = M^*$, and thus $C_T = C^*$, leading to $R_T = R^*$ and $Q_g(a_i) = Q^*_g(a_i)$.

\noindent\textbf{Convergence Rate Analysis.}

To achieve this with confidence level $1 - \delta$, the required number of samples per pair is:
\begin{equation}
T \geq \frac{1}{2 \epsilon^2} \ln\left( \frac{n(n - 1)}{\delta} \right).
\end{equation}
For small $\delta$ and $\epsilon < 0.5$, $T$ scales logarithmically with the number of solutions $n$.

\subsection{Addressing Practical Considerations}

In practice, the true preference probabilities may not be exactly $0$ or $1$ due to noise or overlapping quality scores. To accommodate this:

\begin{itemize}
    \item \textbf{Extended Preference Model}: Assume $P^*(a_i \succ a_j)$ is a strictly increasing function of $\Delta Q^*_{ij} = Q^*(a_i) - Q^*(a_j)$, such as:
    \begin{equation}
    P^*(a_i \succ a_j) = F(\Delta Q^*_{ij}),
    \end{equation}
    where $F$ is a cumulative distribution function (CDF).
    
    \item \textbf{Margin Condition}: Define a margin $m > 0$ such that for all $i \neq j$:
    \begin{equation}
    \left| P^*(a_i \succ a_j) - 0.5 \right| \geq m.
    \end{equation}
    This ensures a minimum separation between preference probabilities.
    
    \item \textbf{Modified Sample Complexity}: With the margin condition, Hoeffding's inequality becomes:
    \begin{equation}
    \mathbb{P}\left( M_T[i, j] \neq M^*[i, j] \right) \leq 2 \exp(-2 T m^2).
    \end{equation}
    To achieve $\mathbb{P}\left( M_T = M^* \right) \geq 1 - \delta$, we need:
    \begin{equation}
    T \geq \frac{1}{2 m^2} \ln\left( \frac{n(n - 1)}{\delta} \right).
    \end{equation}
    
    \item \textbf{Convergence under Noise}: Even with noisy preference probabilities, as long as there is a margin $m > 0$, the convergence of $Q_g(a_i)$ to $Q^*_g(a_i)$ still holds with high probability for sufficiently large $T$.
\end{itemize}

\subsection{Final Conclusion}

We have provided a formal discussion showing that the estimated quantile scores $Q_g(a_i)$, obtained through the EBC method, converge almost surely to the true quantile scores $Q^*_g(a_i)$ as the number of samples $T$ approaches infinity. The finite sample analysis demonstrates that the convergence rate depends logarithmically on the number of solutions and is inversely proportional to the square of the margin $m$ between preference probabilities.


\section{Pseudo-code of main Algorithms}
\begin{algorithm*}[!t]
\SetAlgoLined
\KwIn{Initial state \( s_0 \), search tree \( \mathcal{T} \), max nodes \( N_{max} \), exploration constant \( c \)}
\KwOut{Ranked solution list \( S \)}

Initialize search tree \( \mathcal{T} \) with root node \( s_0 \)\;
\While{number of nodes \( N(\mathcal{T}) < N_{max} \)}{
    \textbf{---------\textit{Selection Phase}---------}\\
    Select a node \( s_i \) which met the dynamic pruning rule from \( \mathcal{T} \) using UCT:
    \[
    a = \arg\max_{a \in A(s)} \left( Q(s, a) + c \cdot \sqrt{\frac{\ln N(s)}{N(s, a)}} \right)
    \]
    
    \textbf{---------\textit{Expansion Phase}---------}\\
    Expand \( s_i \) by generating a successor node \( s' \) using the rewriting process \( R(s_i, c_i) \), where \( c_i = C(s_i) \) is a critique of the current state\;
    Add the new node \( s' \) to \( \mathcal{T} \)\;
    
    \textbf{---------\textit{Evaluation Phase}---------}\\
    
    Compute the value \( Q(s') \) of the new node with Enhanced Borda Count (EBC) method:
    \[
    Q(s') = \alpha Q_g(s') + (1 - \alpha) Q_l(s')
    \]
    where \( Q_g(s') \) is the global value from the win-loss matrix $M$ and \( Q_l(s') \) is the local value from adjacent nodes in \( \mathcal{T} \)\;

    \textbf{---------\textit{Backpropagation Phase}---------}\\
    
    Propagate \( Q(s') \) back to its parent node \( s_i \), updating \( s_i \)'s Q value:
    \[
    Q(s_i) = (1 - \gamma) Q(s_i) + \gamma Q(s')
    \]
    
    \textbf{---------\textit{Check for tree growth limi}t---------}\\
    \If{\( N(\mathcal{T}) \geq N_{max} \)}{
        \textbf{break}\;
    }
}
\Return{Ranked solution list \( S \)}
\caption{Self-Refine applied to Monte Carlo Tree Search (SR-MCTS) Method}
\label{algo:algo1}
\end{algorithm*}

\begin{algorithm*}[!t]
\label{algo:algo2}
\SetAlgoLined
\KwData{$M$ (Binary Reward Matrix),$P$ (Pairwise Preference Reward model)}
\KwResult{$Q$ (quantile rewards), $R$ (ranked node list), $L$ (ranked layers of nodes)}
\DontPrintSemicolon

\SetKwFunction{FEnhancedBordaCount}{EnhancedBordaCount}
\SetKwFunction{FTransitiveClosure}{FillTransitiveClosure}
\SetKwFunction{FRankByWins}{BordaCount}
\SetKwFunction{FCalculateQuantileScores}{CalculateQuantileScores}
\SetKwFunction{FRerank}{Rerank}
\SetKwProg{Fn}{Function}{:}{}

\Fn{\FTransitiveClosure{$M$}}{
    $C \leftarrow \text{all-zero matrix from size of } M$\;
    $C[i,j] \leftarrow -1 \text{ for all } i, j$\;
    $C[i,j] \leftarrow \text{sign}(M[i,j] - 0.5) \textbf{ if } M[i,j] \neq -1$ \textbf{else} $-1$\;
    \For{$k = 0$ \KwTo $|C|-1$}{
        \For{$i = 0$ \KwTo $|C|-1$}{
            \For{$j = 0$ \KwTo $|C|-1$}{
                \If{$C[i,k] = C[k,j]$}{
                    $C[i,j] \leftarrow C[i,k]$
                }
            }
        }
    }
    \KwRet{Updated $C$}\;
}

\Fn{\FRankByWins{$C$}}{
    $D \leftarrow \text{outdegree of each node from } C$\;
    $R \leftarrow \text{argsort}(-D)$\;
    $L$ $\leftarrow$ Define layers  using unique values in $D$\;
    \KwRet{$R$ , $L$ }\;
}

\Fn{\FRerank{$R$, $L$, $C$, $P$}}{
    $G$ $\leftarrow$ Group $R$ by $L$ \;
    $R$ $\leftarrow$ Sort each group in $G$ using local comparisons computed by $P$\;
    $L,C$ $\leftarrow$ Update $L,C$ by $R$\;
     \KwRet{$R$, $L$, Updated $C$}
}

\Fn{\FCalculateQuantileScores{$R$, $L$}}{
    $Q \leftarrow \emptyset$\;
    $S_L \leftarrow \{1 - \frac{l}{\max(L)} : l \in L\}$\;
    $Q[x] \leftarrow S_L[\text{layer of } x]$ for each $x \in R$\;
    \KwRet{$Q$}\;
}

\Fn{\FEnhancedBordaCount{$M$, $q$}}{
    $C$ $\leftarrow$ \FTransitiveClosure{$M$}\;
    $R$, $L$ $\leftarrow$ \FRankByWins{$C$}\;
    $R$, $L$, $C$ $\leftarrow$ \FRerank{$R$, $L$, $C$,$P$}\;
    $Q$ $\leftarrow$ \FCalculateQuantileScores{$R$, $L$}\;
    \KwRet{$Q$, $R$, $L$}\;
}

\caption{Ehanced Borda Count (EBC) Method}
\label{algo:algo2}
\end{algorithm*}
We present the process of Self-Refine applied to Monte
Carlo Tree Search~(SR-MCTS) Method in Algorithm~\ref{algo:algo1} and provide overall pseudo-code for Ehanced Borda Count~(EBC)
Method in Algorithm~\ref{algo:algo2}.

\end{document}